%% file: conf_da.tex
\documentclass[twoside]{article}

\usepackage[accepted]{aistats2020}
\usepackage[round]{natbib}

\usepackage[utf8]{inputenc} % allow utf-8 input
\usepackage[T1]{fontenc}    % use 8-bit T1 fonts
\usepackage{hyperref}       % hyperlinks
\usepackage{url}            % simple URL typesetting
\usepackage{booktabs}       % professional-quality tables
\usepackage{amsfonts}       % blackboard math symbols
\usepackage{nicefrac}       % compact symbols for 1/2, etc.
\usepackage{microtype}      % microtypography

\usepackage{mathtools} % \coloneqq
\usepackage{xcolor}
\usepackage{subcaption}

\input{externals/mylatex/header}

\usepackage{makecell}
\usepackage{tikz}
\usepackage{comment}
\usepackage{multirow}

\providecommand\SP[1]{$\heartsuit$\footnote{SP: {#1}}}

\def\y{\mathbf{{y}}}

\begin{document}

\runningtitle{Calibrated Prediction with Covariate Shift via Unsupervised Domain Adaptation}


\twocolumn[

\aistatstitle{Calibrated Prediction with Covariate Shift \\ via Unsupervised Domain Adaptation}

\aistatsauthor{Sangdon Park \And Osbert Bastani \And James Weimer \And Insup Lee}

%\aistatsaddress{University of Pennsylvania \And University of Pennsylvania \And University of Pennsylvania \And University of Pennsylvania} ]
\aistatsaddress{PRECISE Center \\ University of Pennsylvania} ]

\begin{abstract}
%% Motivation
Reliable uncertainty estimates are an important tool for helping autonomous agents or human decision makers understand and leverage predictive models. 
%Specifically, as the forecaster interplays with other components of systems, the reliability of its label prediction is as important as the label prediction itself. 
%To this end, forecaster calibration methods have been proposed, assuming the training and testing distributions are identical.
%% Challenge
%One approach is to train an uncertainty estimation model that outputs calibrated probabilities for a given classifier.
However, existing approaches to estimating uncertainty largely ignore the possibility of covariate shift---i.e., where the real-world data distribution may differ from the training distribution. As a consequence, existing algorithms can overestimate certainty, possibly yielding a false sense of confidence in the predictive model. We propose an algorithm for calibrating predictions that accounts for the possibility of covariate shift, given labeled examples from the training distribution and unlabeled examples from the real-world distribution.
%However, as a system deploys in the wild, it encounters a new distribution, which differs from the distribution where the forecaster is trained and calibrated. Moreover, the conventional calibration approaches lack to exploit unlabeled examples from the new distribution.
%% our approach
%In this paper, we propose a novel forecaster calibration method that leverages unlabeled examples from the new distribution as well as labeled examples from the original distribution, assuming covariate shift. 
%% key idea: 1) estimate unbiased estimator
Our algorithm uses importance weighting to correct for the shift from the training to the real-world distribution.
%of a labeled example if the example is more likely to see from the new distributions. 
%However, this assumes that the support of original distribution contains that of the new distribution, which seldomly holds. 
%% key idea: 2) enforce assumption
However, importance weighting relies on the training and real-world distributions to be sufficiently close. Building on ideas from domain adaptation, we additionally learn a feature map that tries to equalize these two distributions.
%% results
In an empirical evaluation, we show that our proposed approach outperforms existing approaches to calibrated prediction when there is covariate shift.
\end{abstract}

%% body
\input{sections/intro}
\input{sections/rel}
\input{sections/problem}

%\input{sections/prelim}
\input{sections/approach.tex}

\input{sections/exp}
\input{sections/disc}

%% ACK
\subsubsection*{Acknowledgements}
This work was supported in part by NSF CCF-1910769 and by the Air Force Research Laboratory and the Defense Advanced Research Projects Agency under Contract No. FA8750-18-C-0090.  Any opinions, findings and conclusions or recommendations expressed in this material are those of the author(s) and do not necessarily reflect the views of the Air Force Research Laboratory (AFRL), the Defense Advanced Research Projects Agency (DARPA), the Department of Defense, or the United States Government.

%%
%% references
%%
\bibliography{externals/mybibtex/sml}
\bibliographystyle{abbrvnat}
%\bibliographystyle{plainnat}

\input{sections/appendix.tex}

\end{document}

%% file: externals/mylatex/header.tex
\usepackage{amsmath} % http://ctan.org/pkg/amsmath
\usepackage{bbm} % for mathbbm
\usepackage{amsthm}
\makeatletter
\@ifundefined{proposition}{%
    
}{}
\@ifundefined{definition}{%
}{}
\@ifundefined{lemma}{%
    
}{}
\@ifundefined{corollary}{%
    \newtheorem{corollary}{Corollary}
}{}
\@ifundefined{theorem}{%
    \newtheorem{theorem}{Theorem}
}{}
\@ifundefined{corollary}{%
    
}{}
\@ifundefined{assumption}{%
    
}{}
\@ifundefined{problem}{%
    
}{}
\@ifundefined{problem*}{%
    \newtheorem*{problem*}{Problem}
}{}
\@ifundefined{claim}{%
    
}{}
\makeatother

\newtheorem{innercustomgeneric}{\customgenericname}
\providecommand{\customgenericname}{}
\newcommand{\newcustomtheorem}[2]{%
  \newenvironment{#1}[1]
  {%
   \renewcommand\customgenericname{#2}%
   \renewcommand\theinnercustomgeneric{##1}%
   \innercustomgeneric
  }
  {\endinnercustomgeneric}
}
\newcustomtheorem{customclaim}{Claim}

\providecommand{\eqnref}			[1]		{Equation~(\ref{#1})}
\providecommand{\tabref}			[1]		{Table~\ref{#1}}
\providecommand{\figref}			[1]		{Figure~\ref{#1}}

\providecommand{\eqa}				[1]		{\begin{align}#1\end{align}}
\providecommand{\eqas}			[1]		{\begin{align*}#1\end{align*}}

\providecommand{\ie}{\emph{i.e.,}~}
\providecommand{\eg}{\emph{e.g.,}~}

\providecommand\qcomment[1]{ }

\renewcommand{\(}						{\left(}
\renewcommand{\)}						{\right)}
\renewcommand{\[}						{\left[}
\renewcommand{\]}						{\right]}

\providecommand{\Exp}{\mathbbm{E}}
\providecommand{\Expop}{\mathop{\Exp}}

\def\y{\bm{y}}

\def\fh{\hat{{f}}}
\def\gh{\hat{{g}}}

\def\wh{\hat{{w}}}

\def\As{\mathcal{{A}}}

\def\Ds{\mathcal{{D}}}

\def\Fs{\mathcal{{F}}}
\def\Gs{\mathcal{{G}}}

\def\Ms{\mathcal{{M}}}

\def\Ss{\mathcal{{S}}}
\def\Ts{\mathcal{{T}}}
\def\Us{\mathcal{{U}}}

\def\Ws{\mathcal{{W}}}
\def\Xs{\mathcal{{X}}}
\def\Ys{\mathcal{{Y}}}

%% file: sections/intro.tex
%% !TEX root = ../conf_da.tex

%%
%% Introduction
%%
\section{Introduction}
%% (motivation) reliability of a classifier plays a crucial rule
% nn is an important building block

Reliable uncertainty estimates can substantially improve the usefulness of predictive models. For example, in safety-critical systems such as robotics control, knowing the uncertainty in predictions enables the robot to act more conservatively when uncertainty is higher. Furthermore, in human-in-the-loop decision making, such as healthcare, financial, and legal settings, the uncertainty can help the human decision maker know whether a prediction is trustworthy.
%A neural net forecaster is a crucial building block for a safety-critical system due to its significant progress in various applications (\eg image classification \cite{krizhevsky2012imagenet,lecun1989backpropagation}, object recognition and detection \cite{girshick2014rcnn}, and speech recognition \cite{graves2013speech}). 
% nn outputs a label prediction and a reliability of the prediction, nn needs to satisfy properties
Thus, there has been interest in predictive models that output their uncertainty in the predicted label (\eg object category)~\citep{murphy1972scalar,platt1999probabilistic,zadrozny2001obtaining,zadrozny2002transforming,guo2017calibration,naeini2015obtaining}.
%, needs to satisfy desired safety properties. 
% nn needs to satisfy properties using testing/verification
%To this end, the neural net's \emph{label prediction} can be tested and verified whether it satisfies design-time properties (\eg robustness on a small input perturbations) \cite{gehr2018ai2,katz2017reluplex,pei2017deepxplore,raghunathan2018certified,wong2017provable}, assuming the neural net remains stationary in run-time to maintain the tested and verified properties. 
% nn needs to satisfy properties using calibration

%Moreover, traditionally calibration approaches, which enhance the \emph{reliability} of the label predictions, have been proposed , assuming the training and testing distributions are identical.

%% (motivation) however, the system in the wild suffers a covariate shift

However, work on calibrated prediction has mostly focused on the supervised learning setting, where a labeled validation set can be used to calibrate the predictive model. A major challenge is that oftentimes, the covariate distribution in the training set (the \emph{source distribution}) can be different from the real-world covariate distribution (the \emph{target distribution}), even if the conditional label distribution is the same---this challenge is referred to as \emph{covariate shift}~\citep{shimodaira2000improving}. For example, a robot may be exploring a novel environment, or two hospitals may have different patient populations.
%{\color{red}where the conditional label distributions are the same}.
Having calibrated probabilities that account for covariate shift is important, since understanding uncertainty is particularly useful in settings such as covariate shift that may lead to higher rates of prediction errors~\citep{snoek2019can}. Furthermore, failing to account for uncertainty due to covariate shift can lead an agent to have a false sense of certainty, potentially leading to poor decisions.

%However, one main challenge which the tested, verified, and calibrated neural net faces is that the neural net encounters a covariate-shift when it deploys in the real environment (\ie a covariate-distribution, or a target distribution, when it deploys differs from that the original covariate distribution, or a source distribution), while labeling distributions are the same ).
%% (problem) how to predict the correct confidence under the covariate shift without changing the neural net classifier?
%Under the covariate-shift, we consider a problem where calibrating the reliability of its label prediction while maintaining the tested and verified neural net. 
%% (challenge) leveraging the unlabeled examples from the deploying environment for calibration
%Here, the main challenge is how to leverage the unlabeled examples from the shifted distribution for calibration.

We consider the problem of calibrated prediction when there is covariate shift. In particular, we are interested in the setting of unsupervised domain adaptation, where we have labeled example $(x,y)$ from the source distribution $p$, and unlabeled examples $x$ from the target distribution $q$, where $q$ may differ from $p$.
%% (our approach) auxiliary calibration net
%In this paper, 

Given a classifier $f$, we propose a novel algorithm for training a different model $\fh$ that predicts the uncertainty of the predictions of $f$. We focus on neural networks.
%, which is a neural network that estimates the uncertainty of the the predictions made by $f$.
% by exploiting unlabeled examples from the new environment.
%% bound
The key challenge is how to leverage the unlabeled examples from $q$ to adjust an uncertainty predictor for the supervised learning setting---\eg trained using temperature scaling~\citep{platt1999probabilistic,guo2017calibration}---to account for covariate shift.
%Traditionally, the calibration error is a squared difference between the predicted uncertainty and actual uncertainty.  Then, this error can be upper bounded by the mean-squared classification error, so this upper bound can be used to train a uncertainty estimation model.

We use an approach based on importance weighting.
Our importance weights are estimated based on a source-discriminator that distinguishes whether a given example is from the source or target distributions~\citep{shimodaira2000improving};
%This provides a weighting function where it has a high value if an example is more likely observable in the target distribution, which is called importance weight.
this weighting function increases the uncertainty if an example is likely to have been drawn from the target distribution. Then, we devise a novel upper bound on the expected calibration error consisting of two terms:
% two terms
(i) an importance weighted version of the traditional upper bound, and (ii) the error of the trained source-discriminator. Finally, we propose an algorithm for training an uncertainty prediction model based on minimizing this upper bound.
%we can eventually minimize the expected calibration error with respect to (w.r.t.) the target distribution. 

% assumption
A remaining challenge is that our calibration error bound relies on the assumption that the support of a target distribution is contained within the support of the source distribution, which is a common assumption needed for importance weighting to work \citep{cortes2008sample}. To satisfy this assumption, we use an idea from unsupervised domain adaptation---in particular, we train a feature map such that it is hard to distinguish whether a feature vector represents a source example or a target example~\citep{ben2007analysis}. In particular, the discriminator trained in this approach is in fact a source-discriminator.
%Interestingly, to train the uncertainty estimation model, minimizing our bound and satisfying its assumption is similar to learning an indistinguishable feature map and minimizing weighted classification error, sequentially. 

%% main contributions
Our contributions are: (i) we prove an upper bound on the covariate shifted calibration error (Section~\ref{sec:theory}), (ii) we propose an algorithm for calibrated prediction with covariate shift (Section~\ref{sec:alg}), and
%which decomposed the problem into three parts: (i) distribution-discriminator error minimization, weighted classification error minimization, and indistinguishable feature map learning; 
%(2) we propose a novel calibration method that sqeuantially solves the three decomposed problems; and
(iii) we empirically show that our approach outperforms existing approaches in the presence of covariate shift (Section~\ref{sec:exp}).

%% file: sections/rel.tex
\section{Related Work}

%%
%% calibration
\textbf{Calibration.}
%% overview: goal
The earliest work on calibrated prediction comes from meteorology~\citep{murphy1972scalar,degroot1983comparison}.
%% binary 
In the setting of binary classification, Platt scaling is an effective approach to calibration based on rescaling the probabilities of a classifier using a labeled validation set \citep{platt1999probabilistic}.
%% multi-class: regression
It has been extended to multi-class classification (called temperature scaling)~\citep{guo2017calibration} and to regression~\citep{kuleshov2018accurate}.
%Similar to the Platt scaling, temperature scaling rescales the output score of a multi-class classifier with a positive scalar value \citep{guo2017calibration}. %\osbert{Do you have a citation for this technique?}
%% mutli-cass: binning
Histogram binning is another approach to calibrated prediction~\citep{zadrozny2001obtaining,naeini2015obtaining,zadrozny2002transforming}. 
This approach partitions a classifier score range into bins and assigns calibrated probability to each score bin. 
%% structured prediction
Approaches to calibrated prediction have also been proposed for structured prediction~\citep{kuleshov2015calibrated}. 
%% confidence set
Finally, approaches that estimate a confidence set, rather than calibrated prediction, have been proposed~\citep{papadopoulos2008inductive,vovk2013conditional,barber2019predictive,Park2020PAC}.
%% difference
All of these techniques are focused on the supervised learning setting, and do not account for covariate shift.
%The conventional calibration method assumes the source and target distributions are identical (iid assumption), which can be violated in the operational time of a system with a classifier. 
%% other setup
There has been some recent work on calibrated prediction under covariate shift~\citep{kuleshov2017estimating,snoek2019can}; however, these approaches target different settings than ours---e.g., they require labels from the target domain~\citep{kuleshov2017estimating}. 
%r do not leverage unlabeled examples from the target domain at all~\citep{}.
%\SP{{\color{red}I replaced the CAM paper to a similar NeurIPS 19 paper \citep{snoek2019can}. Feel free to drop it if it's improper.}}
%these approaches do not leverage the unlabeled data from the target distribution \citep{cam19draft}, or .
%, so their uncertainty estimates are not tailored to the observed data distribution.

%%
%% DA
\textbf{Domain adaptation.}
The goal of unsupervised domain adaptation is to transfer a classifier trained on a source distribution $p$ to a target distribution $q$---e.g., $q$ may be a covariate shifted version of $p$. 
%% importance weighting
Importance weighting is one approach to domain adaptation~\citep{bickel2007discriminative,huang2007correcting,kanamori2009least,shimodaira2000improving,sugiyama2008direct}. This approach reweights the labeled training data from the source domain by the \emph{importance weight} $w(x)=\frac{q(x)}{p(x)}$, where $p$ and $q$ are source and target distributions, respectively. Intuitively, this reweighting upweights samples from $p$ that are rare according to $p$ but common according to $q$. In particular, the importance weighted empirical risk on the source distribution is unbiased estimate of the empirical risk on the target distribution \citep{cortes2008sample}. 
%% 
%The key of this approach is estimating the importance weight, but generally it is know that estimating $p$ and $q$ individually is hard.
A key challenge is how to estimate the importance weight~\citep{bickel2007discriminative,huang2007correcting,kanamori2009least,sugiyama2008direct}. Our work builds on the ideas from~\cite{bickel2007discriminative}.
%% weakness of IW

An important assumption on importance weighting is that the support of the target distribution $q$ is included in the support of the source distribution $p$~\citep{cortes2008sample}; furthermore, these algorithms perform poorly when the importance weights $w(x)$ become large---\ie if $p(x)$ is small where $q(x)$ is not.

%% feature learning approach
An alternative approach called \emph{indistinguishable feature learning}
%\SP{{\color{red}does this mean ``adversarial training''? Since we mentioned that IW is one approach to domain adaptation.}}
has been proposed to address these issues~\citep{ben2007analysis,JMLR:v17:15-239,hoffman2017cycada,long2017deep,long2018conditional,sankaranarayanan2018generate}. The key idea is to learn an \emph{indistinguishable feature map}, which maps $x$ to a latent space, such that in the latent space, the source distribution is close to the target distribution. Then, labels from the source distribution should be transferable to the target distribution. Our approach leverages this approach to avoid the potential blowup in $w(x)$.

%% file: sections/problem.tex
\section{Problem Formulation}

%% overview
We introduce the calibrated prediction problem in the supervised setting, and then state our problem of calibrated prediction under the covariate shift.
%% notations

%%
%% calibration
\textbf{Calibrated prediction for supervised learning.}
Let $\Xs$ be the space of covariates, 
$\Ys = \{1, 2\, \dots, K \}$ be the set of labels. A \emph{forecaster} $f: \Xs \rightarrow [0, 1]^{|\Ys|}$ returns a distribution over labels (\ie $\sum_{y\in\Ys} f(x)_y = 1$ for any $x \in \Xs$),
%% overview: motivation, illustrative example
where $f(x)_y$ is an estimate of the probability of label $y\in\Ys$.
%and the probabilities are generally exploited in other system components for decision making. 
%The estimated probabilities by the forecaster are required to be reliable.
We let $\Fs$ denote the space of forecasters.
%Intuitively, the estimated probabilities are calibrated if, for all of the examples $x$ for which $f(x)_y=c$, exactly $c$ fraction of them have true label equal to $y$.
%For example, given all days when the forecaster said it is rainy with $90\%$ confidence, the ratio of rainy days should be $0.9$ to be called as a calibrated forecaster.
%% ``calibrated'' definition in binary
A forecaster $f: \Xs \rightarrow [0, 1]^{|\Ys|}$ is \emph{well-calibrated} with respect to a distribution $p$ over  $\Xs \times \Ys$ if
\begin{align}
\text{Pr}_{p(x,y)}( y = k \mid f(x)_k = t ) = t \label{def:calibrated}
\end{align}
for all $k \in \Ys$ and $t \in [0, 1]$~\citep{degroot1983comparison,zadrozny2002transforming}.
%% interpretation 
%Based on this definition, if the well-calibrated forecaster says an observed example $x$ is likely to be label $1$ with probability $c$ (\ie $f(x)=c$), then $100c\%$ examples among examples where the forecaster said label $1$ with probability $c$ (\ie $\{x \;|\; f(x)=c\}$) have the true label $1$. 
%In this sense, we call the well-calibrated forecaster \emph{reliable}.
%% definition: multiclass extension
%% note: decision rule agnostic
%Note that this definition of calibration is agnostic to the decision-rule---\ie a calibrated forecaster can be used with a deterministic decision-rule (\ie $\yh = \arg\max_{k \in \Ys} f_k(x)$ where $\yh$ is a label prediction) or a randomized one (\ie $\yh$ drawn from $f(x)$). This property is different from the definition of calibration in \cite{guo2017calibration}, which depends on a decision-rule.
%\SP{Do I need this sentence? we are given with a classifier, so need a decision-rule dependent definition?  }
%\SP{Note that we derive a calibration method based on the decision-rule agnostic definition, but we decision-rule dependent one to evaluate a calibrator...}
%% how to find a (well-)calibrated forecaster?
%% 1. minimize the calibration error: || f(x) - c(x) ||. (-) a reliable forecaster is not necessarily useful.
Define $c:\Xs\to[0,1]^{|\Ys|}$ by
\begin{align*}
c(x')_k &= \text{Pr}_{p(x,y)}( y = k \mid f(x)_k = f(x')_k ) \\
&= \Expop_{p(x,\y)}[ \y_k \mid f(x')_k ],
\end{align*}
where $\y \in \{0, 1\}^{|\Ys|}$ is a one-hot encoding of the label $y \in \Ys$ (\ie $\y_y = 1$, and $\y_{y'} = 0$ for $y'\neq y$), and $c$ is implicitly a function of $f$. It can be shown that we can calibrate $f$ by minimizing
%\begin{align*}
$\Expop_{p(x)} [ \| f(x) - c(x)  \|^{2} ]$,
%\end{align*}
called \emph{calibration error}~\citep{murphy1972scalar}.% limitation

Having small calibration error is not sufficient~\citep{murphy1977reliability,kuleshov2015calibrated}---\eg a forecaster that always returns the base rate of each label without looking at a given example (\ie $f(x) = \Expop_{p(\y)}[ \y ]$) is well calibrated since $c(x) = \Expop_{p(\y)}[ \y ]$, but is not useful since its predictions do not depend on $x$.
%% 2. minimize the classification error: || f(x) - y ||. To achieve both reliability and accuracy.
Intuitively, we want a forecaster that predicts the right label in addition to being calibrated.

Consider the following decomposition of the expected mean-squared classification error~\citep{murphy1972scalar}:
\eqa{
	\underbrace{\Expop \! \[ \| f(x) - \y \|^{2} \]}_\text{{classification error}}  
		\! = \! \underbrace{\Expop \! \[ \| f(x) -  c(x)  \|^{2} \]}_{\text{calibration error}} \!+ 
		1 \! - \! \underbrace{\Expop \! \[ \|c(x)\|^{2} \]}_{\text{sharpness}}, \!\! \label{eq:conv_calibration_goal}
}
where we have dropped $p(x,\y)$ since it is clear from context.
%% interpertation
The last term is the \emph{sharpness} of a forecaster, which intuitively rewards the forecaster for outputting probabilities closer to 0 or 1. In particular, a forecaster that achieves low classification error is both calibrated and useful.
%% emphasize ``supervised'' calibration
Thus, one approach to learning a calibrated and useful forecaster is to minimize (\ref{eq:conv_calibration_goal}) on a validation set of labeled examples.
\footnote{The mean-squared classification error is also called a multi-class extension of the Brier score \citep{brier1950verification}; there are other calibration approaches that minimize different losses than the mean-squared error---\eg \citep{platt1999probabilistic, guo2017calibration}.}
If $\Fs$ has low VC dimension, then the predicted uncertainties will generalize well \citep{vapnik1995nature}.

A common approach, called \emph{temperature scaling}~\citep{platt1999probabilistic,guo2017calibration}, is to first learn a model $f$ (\eg a neural network) $f$ with high VC dimension, and then calibrate by rescaling its predicted uncertainties using just a single parameter called the \emph{temperature}. We describe this approach in detail in Section~\ref{subsec:alg}. Note that this approach assumes the test distribution equals the training distribution---thus, it cannot account for the potential for increased uncertainty in the predictions in the presence of covariate shift.

%%
%% Problem
\textbf{Calibrated prediction with covariate shift.}
%\SP{{\color{red}do we want to use $x_s, y_s$ for a source labeled example and $x_t$ for the target example, which a reviewer suggested?}}
%% overview
We consider calibrated probabilities that account for a covariate shift from a source distribution $p(x,y)=p(y\mid x)\cdot p(x)$ (the training distribution) to a target distribution $q(x,y)=q(y\mid x)\cdot q(x)$ (the test distribution), in the setting of unsupervised domain adaptation.
%% define ``covariate shift'', emphasize that this assumption usually holds
%The covariate shift describes a dataset shift between a source distribution $p(x, y)$ and a target distribution $q(x, y)$ where 
We make the standard assumption that the source and target distributions may be different (\ie we may have $p(x) \neq q(x)$) but
the labeling distributions are identical (\ie $p(y\mid x) = q(y\mid x)$) \citep{shimodaira2000improving}. 
%% unsupervised

Then, our goal is to learn a forecaster that is calibrated to the target distribution $q(x, y)$, given labeled examples from the source distribution and unlabeled examples from the target distribution. More precisely, given (i) labeled examples $(x,y)\sim p$ from the source distribution, and (ii) unlabeled examples $x\sim q$ from the target distribution, our goal is to learn a forecaster $\fh \in \Fs$ that minimizes a combination of the expected calibration error and sharpness with respect to the target distribution $q$--\ie
\eqa{
	\min_{\fh \in \Fs} \Expop_q\[ \| \fh(x) - c(x) \|^{2} \] + 1 - \Expop_q \[ \|c(x)\|^{2} \],
	\label{eq:goal}
}
where $c(x) = \Expop_q[ \y \mid \fh(x) ]$, and $q(y \mid x)=p(y \mid x)$ is the (unknown) labeling distribution.

Alternatively, we can also consider the \emph{recalibration problem}~\citep{kuleshov2017estimating}. In this setting, we are given a classifier $f:\Xs\to\Ys$. Then, our goal is to learn a forecaster that estimates the uncertainty of the predictions made by $f$---\ie,
%\begin{align*}
%\min_{\fh \in \Fs} \Expop_{q}\left[\| \fh(x) - c_f(x) \|^{2}\right] + 1 - \Expop_q \left[ \|c_f(x)\|^{2} \right],
%\end{align*}
%where $c_f(x) = \Expop_q[ \y \mid f(x)]$, $q(y\mid x)=p(y\mid x)$ as before, 
\begin{align*}
\min_{\fh \in \Fs} \Expop_{q}\left[ | \fh(x)_{f(x)} - c(x)_{f(x)} |^{2}\right] + 1 - \Expop_q \left[ \|c(x)\|^{2} \right],
\end{align*}
where $c(x) = \Expop_q[ \y \mid \fh(x) ]$, and $q(y \mid x)=p(y \mid x)$ as before. Our techniques can be applied both to calibration and to recalibration.
%\SP{{\color{blue}
%The following could be the recalibration formulation, which aligns with our usage:
%\begin{align*}
%\min_{\fh \in \Fs} \Expop_{q}\left[ | \fh(x)_{f(x)} - c(x)_{f(x)} |^{2}\right] + 1 - \Expop_q \left[ \|c(x)\|^{2} \right], (100)
%\end{align*}
%where $c(x) = \Expop_q[ \y \mid \fh(x) ]$, and $q(y \mid x)=p(y \mid x)$ as before. Here, ``aligns with our usage'' means, 1) it aligns with the evaluation metric and 2) our approach, where we just solve calibration for recalibration, i.e., minimizing the classification error without considering the given classifier. I think this can be justified since (3) is the upper bound of (100).}}

%
%% problem statement

% connection this forecaster and a given classifier
%Note that we design a forecaster such that it embeds a property of a given classifier (\eg sharing a feature map) and outputs a reliability of the classifier on the behalf of it.

%% file: sections/approach.tex
%% !TEX root = ../conf_da.tex

%%
%% Approach
%% 

%%
%%
%%
%\section{Algorithm for Calibrated Prediction with Covariate Shift}
\section{Upper Bound on Covariate Shifted Calibration Error}
\label{sec:theory}

%% overview: main result + highlight importance weighting + highlight domain-discriminator learning + algorithm
%First, we state our main result on the upper bound, followed by the two key ideas, (i) importance weighting and (ii) source-discriminator learning, with related proofs. Finally, we propose a calibration method that minimizes the upper bound. 
%%
%%
%% 4. (ours 1) we calibrate a forecaster with labeled examples from the source distributions by weighting covariate-shift ratio
Our algorithm for learning a calibrated forecaster in the setting of covariate shift is based on minimizing a novel upper bound on expected calibration error with respect to the target distribution. In this section, we describe the upper bound. This bound accounts for the mismatch between source and target distributions by using importance weighting---i.e., it weighs each labeled example from the source distribution by the \emph{importance weight} $w(x)=\frac{q(x)}{p(x)}$ (Section~\ref{subsec:IW}).
%% 5. (ours 2) we estimate the covariate-shift ratio by a domain-discriminator classifier. 
The importance weight is a priori unknown; thus, we estimate it using a source-discriminator (Section~\ref{subsec:domain_discriminator}).
%% (ours 3) indistinguishable feature learning

\subsection{Assumptions}

%% (provides intuition) how to to leverage labeled and unlabeled examples for calibration?
Recall that to find a forecaster that minimizes the calibration error and maximizes sharpness (\ref{eq:goal}), we can minimize the mean-squared classification error with respect to the target distribution:
\begin{align*}
\Expop_{q} \[ \| \fh(x) - c(x) \|^{2} \] \leq \Expop_{q} \[ \| \fh(x) - \y \|^{2} \].
\end{align*}
%% 1. (challenge) no label on examples from target distribution
A key challenge to this approach is that we do not have labeled examples from the target distribution.
%% 2. (assumption) we assume we've seen the meaningful number of example from the source distribution (ratio is positive and bounded)
To address this challenge, we make the standard \emph{bounded importance weight} assumption, which says that examples with high probability according to the target distribution have non-negligible probability according to the source distribution~\citep{bickel2007discriminative,huang2007correcting, kanamori2009least,shimodaira2000improving,sugiyama2008direct}---i.e., we assume that
\begin{align*}
0 \leq \frac{q(x)}{p(x)} \leq U \hspace{0.2in} (\forall x ~\text{s.t.}~ p(x) \neq 0)
\end{align*}
%% 3. (assumption) we assume the labeling distribution does not change (covariate shift)
for some $U>0$. Moreover, we make the standard \emph{covariate shift} assumption, which says the source and target label distributions are identical---i.e.,
$p(y\mid x)=q(y\mid x)$ for all $x$ such that $p(x)\neq0$ and $q(x)\neq0$ and $y \in \Ys$. This assumption is needed for us to leverage the labeled examples from the source distribution.

\subsection{Importance Weighting} \label{subsec:IW}

%% 1. reformulate the objective in an importance weighting style
The expected mean-squared classification error with respect to the target distribution is decomposed into two terms: (i) a term for the expected mean-squared classification error weighted by the importance weight, and 
(ii) a term for the importance weight estimation error. In particular, we have
\begin{align}
	&\Expop_{q} \[ \| \fh(x) - \y \|^{2} \] \nonumber \\
	&= \Expop_{p} \[ \| \fh(x) - \y \|^{2} w(x) \] \nonumber \\	
		&= \Expop_{p} \[ \| \fh(x) - \y \|^{2} ( \wh(x) + w(x) - \wh(x) ) \] \nonumber \\	
		&= \! \Expop_{p} \! \[  \| \fh(x) \!-\! \y \|^{2} \wh(x) \] \!\!+\!  \Expop_{p} \! \[ \| \fh(x) \!-\! \y \|^{2} ( w(x) \!-\! \wh(x) ) \]\!, \label{eq:W_1_2}
\end{align}
where the first equality holds due to our covariate shift assumption, $w(x) = \frac{q(x)}{p(x)}$ is the importance weight, and $\wh$ is an estimate of $w$.
%% interpretation
The first term of (\ref{eq:W_1_2}) is the conventional expected mean-squared classification error with respect to the source distribution, but each example in the mean-squared error is reweighted by its importance weight.
%% 2. upper bound the second weighting estimation error term
Next, the second term in (\ref{eq:W_1_2}) is bounded as follows:
\begin{align}
	&\Expop_{p} \[ \| \fh(x) - \y \|^{2} ( w(x) - \wh(x) ) \] \nonumber \\
		&\leq \sqrt{ \Expop_{p} \[ \| \fh(x) - \y \|^{4} \] \Expop_{p} \[ ( w(x) - \wh(x) )^{2} \]} \nonumber \\
		&\leq \frac{1}{2} \( \Expop_{p} \[ \| \fh(x) - \y \|^{4} \] +  \Expop_{p} \[ ( w(x) - \wh(x) )^{2} \] \) \nonumber \\
		&\leq \frac{1}{2} \( \Expop_{p} \[ \| \fh(x) - \y \|^{2} \] +  \Expop_{p} \[ ( w(x) - \wh(x) )^{2} \] \), \label{eq:W_2_3}
\end{align}
where the first inequality is due to Cachy-Schwarz, the second is due to the arithmetic-mean geometric-mean inequality, and the third holds since $ \| \fh(x) - \y \|^{2} \leq 1$ for any $\fh$, $x$ and $\y$.
% tight
Note that the bound in (\ref{eq:W_2_3}) is tight when $\fh(x) = \y$ and $w(x) = \wh(x)$.

%% interpretation
The first term in (\ref{eq:W_2_3}) is the usual expected mean-squared classification error.
%which will be combined later with \eqnref{eq:W_1_2}. 
The second term is the expected mean-squared importance weight estimation error with respect to the source distribution. Intuitively, finding a good estimate of the importance weight is crucial since the estimate is used to reweight classification error in (\ref{eq:W_1_2}). We describe how we use a source-discriminator to estimate this quantity in the following section.

\subsection{Source-Discriminators} \label{subsec:domain_discriminator}

A \emph{source-discriminator} $\gh:\Xs\to[0,1]$ is a model that predicts whether an example $x\in\Xs$ was sampled from the source distribution $p$ or the target distribution $q$. We use $\Gs$ to denote a space of source-discriminators.
%% overview: covariate ratio <-> classifying domain
Following~\cite{bickel2007discriminative}, we can use a source-discriminator to predict the importance weight $w(x)$.
%% setup for domain classification

First, consider the distribution $r$ over $(x,s)\in\Xs\times\{0, 1\}$ defined by $r(s)=0.5$ for each $s\in\{0,1\}$ and
\begin{align*}
r(x \mid s) &= \begin{cases}p(x) &\text{if}~s=1 \\ q(x)&\text{otherwise}.\end{cases}
\end{align*}
In other words, we sample $(x,s)\sim r$ as follows: (i) sample $s\sim\text{Bernoulli}(0.5)$, and (ii) sample $x\sim p$ if $s=1$ and $x\sim q$ if $s=0$.
%% classifier <-> ratio
Then, we have
\begin{align*}
w(x) = \frac{r(x \mid s=0)}{r(x \mid s=1)}.
\end{align*}
The optimal source-discriminator is
\begin{align*}
g(x) = r(s=1 \mid x).
\end{align*}
Thus, we can express the importance weight $w(x)$ in terms of $g(x)$ as follows~\citep{bickel2007discriminative}:
\begin{align*}
	g(x) &= r(s=1 \;|\; x)
		%&= \frac{r(x \;|\; s=1) r(s=1)}{r(x \;|\; s=1) r(s=1) + r(x \;|\; s=0) r(s=0)} \\
		%&= \frac{r(x \;|\; s=1)}{r(x \;|\; s=1) + r(x \;|\; s=0)} \\
		= \frac{1}{1 + w(x)}.
\end{align*}
The last equality follows by Bayes' theorem. By our assumption that $w(x)\le U$, we have
\begin{align}
\label{eqn:gbound}
\frac{1}{1+U} \leq g(x) \leq 1.
\end{align}
Given unlabeled data from $p$ and $q$, we can learn an estimate $\gh$ of $g$. We enforce that (\ref{eqn:gbound}) holds for $\gh$ as well.
%% connection to the previous eq
Thus, the second importance weight estimation error term in (\ref{eq:W_2_3}) can be rewritten as follows:
\begin{align}
	&\Expop_{p} \[ \( w(x) - \wh(x) \)^{2} \] \nonumber \\
	&= \Expop_{p} \[ \( \frac{g(x) - \gh(x)}{g(x) \gh(x)} \)^{2} \] \nonumber \\
		&\leq (1+U)^{4} \Expop_{p} \[ \( g(x) - \gh(x)\)^{2} \] \nonumber \\
		&= (1+U)^{4} \Expop_{r} \[ \( g(x) - \gh(x)\)^{2} \frac{p(x)}{r(x)} \] \nonumber \\
		&\leq 2 (1+U)^{4} \Expop_{r} \[ \( g(x) - \gh(x)\)^{2} \] \nonumber \\
		&= 2 (1+U)^{4} \! \[ \Expop_{r} \[ \( s - \gh(x)\)^{2} \] \! - \! \Expop_{r} \[ \( s - g(x)\)^{2} \] \]. \!\! \label{eq:DD_2_3}
\end{align}
The first inequality holds since $g(x)^{-1}$ and $\gh(x)^{-1}$ are bounded by $1+U$, the second holds since $p(x) \leq 2r(x)$ by the definition of $r(x)$,
% the last important equation to make our bound tight
and the last holds since
\eqas{
	&\Expop_{r} \[ \( s - \gh(x)\)^{2} \] \\
	&= \Expop_{r} \[ \( s - g(x) + g(x) - \gh(x) \)^{2} \] \\
	&= \Expop_{r} \[ \( s - g(x) \)^{2} \] + \Expop_{r} \[ \( g(x) - \gh(x) \)^{2} \] \\
	&\hspace{0.2in} + 2  \Expop_{r} \[ (s - g(x))(g(x) - \gh(x)) \] \\
	&= \Expop_{r} \[ \( s - g(x) \)^{2} \] + \Expop_{r} \[ \( g(x) - \gh(x) \)^{2} \],
}
where the last equality holds since $\Expop_r \[ s \;|\; x \] = g(x)$.
% tight
Note that if $g(x) = \gh(x)$, the bound (\ref{eq:DD_2_3}) is tight.

\subsection{Main Result}

%% summary

Combining (\ref{eq:W_1_2}), (\ref{eq:W_2_3}), and (\ref{eq:DD_2_3}) yields the following upper bound on the expected classification error:
%% main theorem
\begin{theorem} \label{thm:main}
%Suppose our assumptions hold.
For any forecaster $\fh \in \Fs$ and any source-discriminator $\gh \in \Gs$ satisfying (\ref{eqn:gbound}),
%a source distribution $p$ over $\Xs \times \Ys$ where $p(x) \neq 0$ for any $x \in \Xs$,
%source distribution $p$ over $\Xs \times \Ys$, and
%a target distribution $q$ over $\Xs \times \Ys$ such that $\frac{q(x)}{p(x)} \leq U$ and $q(y \;|\; x)= p(y \;|\; x)$ for any $x, y \in \Xs \times \Ys$,
%target distribution $q$ over $\Xs \times \Ys$, if $0 \leq \frac{q(x)}{p(x)} \leq U$ and $q(y \mid x)= p(y \mid x)$ for any $x \in \{ x \mid p(x) \neq 0 \text{~and~} q(x) \neq 0 \}$ and  $y \in \Ys$,
%the following inequality holds:
%% improved version
we have
\eqa{
	&\Expop_{q} \[ \| \fh(x) - c(x) \|^{2} \] \leq - %\delta(\lambda, r, g)
	\lambda \Expop_{r} \[ \( g(x) - s \)^{2} \]
	\nonumber \\
		&\!+\! \underbrace{\Expop_{p} \! \[ \| \fh(x) - \y \|^{2} \! \( \frac{1}{\gh(x)} \! - \! \frac{1}{2} \) \]}_{\text{weighted classification error}}
		\!+ \lambda \!\! \underbrace{\Expop_{r} \[ \( \gh(x) \! - \! s \)^{2}  \]}_{\text{\makecell{source-discriminator \\ error}}} \!\!\!\!, \!\! \label{eq:main_bound}
}
where 
$c(x) = \Expop_q[ \y \mid \fh(x) ]$
%$r(x \;|\; s = 1) = p(x)$, 
%$r(x \;|\; s = 0) = q(x)$, 
%$r(s=1) = r(s=0) = 0.5$,
and $\lambda = (1 + U)^{4}$.
%and
%$\delta(\lambda, r, g) = \lambda \Expop_{r} \[ \( g(x) - s \)^{2} \]$.
%\begin{comment}
%{\color{red}
%\eqa{
%	& \Expop_{q} \[ \| \fh(x) - c(x) \|^{2} \]
%		\leq \underbrace{\Expop_{p} \[ \| \fh(x) - \y \|^{2} \( \frac{1}{\gh(x)} - \frac{1}{2} \) \]}_{\text{weighted classification error}} \nonumber \\
%		&+ \underbrace{ \Expop_{p} \[ \( \frac{1 - \gh(x)}{\gh(x)} \)^{2} \] -2 \Expop_{q} \[ \frac{1 - \gh(x)}{\gh(x)} \] + \delta(p, q)}_{\text{source-discriminator error}}, \label{eq:main_bound}
%} 
%where 
%$c(x) = \Expop_q[ \y \mid \fh(x) ]$,
%and
%$\delta(p, q) = \Exp_q\[ \nicefrac{q(x)}{p(x)} \]$.
%}
%\end{comment}
\end{theorem}
%\SP{{\color{red}do we need a single proof box in the appendix? a reviewer mentioned that ``Theorem 1 can be presented more formally with a coherent proof.''}}
%% interpretation.
In (\ref{eq:main_bound}), the second term is the importance weighted classification error, where the importance weight is using the given source-discriminator. The first and third term are the source-discriminator estimation error. This result extends the one in \cite{cortes2010learning} to the case where the importance weights are unknown and must be estimated from unlabeled data.
%% tightness
Note that when $\fh(x) = \y$ and $\gh(x) = g(x)$, the bound (\ref{eq:main_bound}) is tight.

%% related work
%Compared to \cite{bickel2007discriminative}, we explicitly consider feature learning procedure to satisfy the bounded importance weight assumption. 

%% conclusion

%{\color{red}
%\eqa{
%	\Expop_{p} \[ \( w(x) - \wh(x) \)^{2} \] 
%		&= \Expop_{p} \[ \( \frac{1}{g(x)} - \frac{1}{\gh(x)} \)^{2} \] \\
%		&= \Expop_{r} \[ \( \frac{1}{g(x)} - \frac{1}{\gh(x)} \)^{2} \frac{p(x)}{r(x)} \] \\
%		&\leq 2 \Expop_{r} \[ \( \frac{1}{g(x)} - \frac{1}{\gh(x)} \)^{2} \] \\
%		&\leq 2 \Expop_{r} \[ \( \frac{1}{y} - \frac{1}{\gh(x)} \)^{2} \],
%}
%$y \in \{ \frac{1}{1+U},  1 \}$
%}

%%
%%
\section{Algorithm}
\label{sec:alg}

%% calibration net
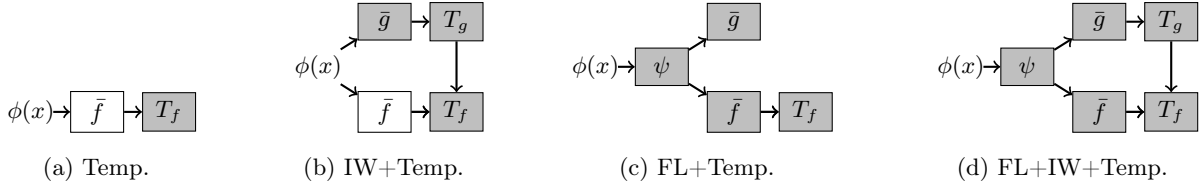
\begin{figure*}[t!]
\centering
\small
%%temp
\hspace{-20pt}
\begin{subfigure}[b]{0.2\textwidth}
\centering
\begin{tikzpicture}[ampersand replacement=\&, scale=1.0]
	%% setups
	\tikzstyle{myblock} = [draw, rectangle,minimum height=1.5em,minimum width=1.5em, text width=1.5em, align=center]
	\tikzstyle{mytext} = [rectangle,thick,minimum height=1.5em,minimum width=1.5em, text width=1.5em, align=center]
	\tikzstyle{myconnector} = [->,thick]

	%% blocks
	\matrix [column sep=1.75ex, row sep=2ex] {
		\node[mytext] (input) {$\phi(x)$}; 
		\&
		\node[myblock] (fbar) {$\bar{f}$}; 
		\&
		\node[myblock,fill=gray!50] (Tf) {$T_f$}; 
		\\
	};
	\draw[myconnector] (input) -- (fbar);
	\draw[myconnector] (fbar) -- (Tf);
\end{tikzpicture}
\caption{Temp.}
\label{fig:cal_net_1}
\end{subfigure}
%%Temp+IW
\hspace{5pt}
\begin{subfigure}[b]{0.2\textwidth}
\centering
\begin{tikzpicture}[ampersand replacement=\&, scale=1.0]
	%% setups
	\tikzstyle{myblock} = [draw, rectangle,minimum height=1.5em,minimum width=1.5em, text width=1.5em, align=center]
	\tikzstyle{mytext} = [rectangle,thick,minimum height=1.5em,minimum width=1.5em, text width=1.5em, align=center]
	\tikzstyle{myconnector} = [->,thick]

	%% blocks
	\matrix [column sep=1.75ex, row sep=2ex] {
		\node[mytext] (input) {$\phi(x)$}; \&
		\node[myblock,fill=gray!50, yshift=0.6cm] (gbar) {$\bar{g}$}; 
		\node[myblock, yshift=-0.6cm] (fbar) {$\bar{f}$}; 
		\&
		\node[myblock,fill=gray!50, yshift=0.6cm] (Tg) {$T_g$}; 
		\node[myblock,fill=gray!50, yshift=-0.6cm] (Tf) {$T_f$}; 
		\\
	};
	\draw[myconnector] (input) -- (gbar);
	\draw[myconnector] (input) -- (fbar);
	\draw[myconnector] (gbar) -- (Tg);
	\draw[myconnector] (fbar) -- (Tf);
	\draw[myconnector] (fbar) -- (Tf);
	\draw[myconnector] (Tg) -- (Tf);
\end{tikzpicture}
\caption{IW+Temp.}
\label{fig:cal_net_2}
\end{subfigure}
%%Temp+FN
\hspace{5pt}
\begin{subfigure}[b]{0.24\linewidth}
\centering
\begin{tikzpicture}[ampersand replacement=\&, scale=1.0]
	%% setups
	\tikzstyle{myblock} = [draw, rectangle,minimum height=1.5em,minimum width=1.5em, text width=1.5em, align=center]
	\tikzstyle{mytext} = [rectangle,thick,minimum height=1.5em,minimum width=1.5em, text width=1.5em, align=center]
	\tikzstyle{myconnector} = [->,thick]

	%% blocks
	\matrix [column sep=1.75ex, row sep=2ex] {
		\node[mytext] (input) {$\phi(x)$}; \&
		\node[myblock,fill=gray!50] (feature) {$\psi$}; \&
		\node[myblock,fill=gray!50, yshift=0.6cm] (gbar) {$\bar{g}$}; 
		\node[myblock,fill=gray!50, yshift=-0.6cm] (fbar) {$\bar{f}$}; 
		\&
		\node[myblock,fill=gray!50, yshift=-0.6cm] (Tf) {$T_f$}; 
		\\
	};
	\draw[myconnector] (input) -- (feature);
	\draw[myconnector] (feature) -- (gbar);
	\draw[myconnector] (feature) -- (fbar);
	\draw[myconnector] (fbar) -- (Tf);
	\draw[myconnector] (fbar) -- (Tf);
\end{tikzpicture}
\caption{FL+Temp.}
\label{fig:cal_net_3}
\end{subfigure}
%%Temp+FL+IW
\hspace{15pt}
\begin{subfigure}[b]{0.24\linewidth}
\centering
\begin{tikzpicture}[ampersand replacement=\&, scale=1.0]
	%% setups
	\tikzstyle{myblock} = [draw, rectangle,minimum height=1.5em,minimum width=1.5em, text width=1.5em, align=center]
	\tikzstyle{mytext} = [rectangle,thick,minimum height=1.5em,minimum width=1.5em, text width=1.5em, align=center]
	\tikzstyle{myconnector} = [->,thick]

	%% blocks
	\matrix [column sep=1.75ex, row sep=2ex] {
		\node[mytext] (input) {$\phi(x)$}; \&
		\node[myblock,fill=gray!50] (feature) {$\psi$}; \&
		\node[myblock,fill=gray!50, yshift=0.6cm] (gbar) {$\bar{g}$}; 
		\node[myblock,fill=gray!50, yshift=-0.6cm] (fbar) {$\bar{f}$}; 
		\&
		\node[myblock,fill=gray!50, yshift=0.6cm] (Tg) {$T_g$}; 
		\node[myblock,fill=gray!50, yshift=-0.6cm] (Tf) {$T_f$}; 
		\\
	};
	\draw[myconnector] (input) -- (feature);
	\draw[myconnector] (feature) -- (gbar);
	\draw[myconnector] (feature) -- (fbar);
	\draw[myconnector] (gbar) -- (Tg);
	\draw[myconnector] (fbar) -- (Tf);
	\draw[myconnector] (fbar) -- (Tf);
	\draw[myconnector] (Tg) -- (Tf);
\end{tikzpicture}
\caption{FL+IW+Temp.}
\label{fig:cal_net_4}
\end{subfigure}
\caption{Calibration algorithms. A gray block means the corresponding layer should be learned.
%using a training set (without target labels) and a validation set (i.e., learned by our calibration algorithm).
Here, $\phi$ is a feature map from a given classifier $f$ to be calibrated, $T_g \bar{g}$ is a learned source-discriminator, and $\psi$ is a learned indistinguishable feature representation. (a) Learn a forecaster $T_f \bar{f} \circ \phi$ using temperature scaling (Temp.).
%approach to calibrated prediction for the supervised setting
~\citep{guo2017calibration}. (b) Learn a forecaster $T_f \bar{f} \circ \phi$ using importance weighting (IW) in addition to Temp., but not feature learning (FL). (c) Learn a forecaster $T_f \bar{f} \circ \psi \circ \phi$ using Temp. and FL, but not IW. (d) Our full algorithm; learn a forecaster $T_f \bar{f} \circ \psi \circ \phi$ using Temp., IW, and FL.}
\label{fig:cal_nets}
\end{figure*}

%% overview

Our algorithm uses the upper bound (\ref{eq:main_bound}) in conjunction with the temperature scaling approach to calibrated prediction~\citep{guo2017calibration} (Section~\ref{subsec:alg}). However, this approach depends on the standard assumption that the importance weights are bounded, which may not always be satisfied. As a heuristic to encourage the satisfaction of this assumption, we build on ideas from domain adaptation---our algorithm learns a \emph{feature map} such that the feature distributions induced from the source and target distributions are indistinguishable (Section~\ref{subsec:algind}).

\subsection{Temperature Scaling}
\label{subsec:alg}

We build on the idea of temperature scaling to perform calibrated prediction~\citep{platt1999probabilistic}. At a high level, this approach takes as given an uncalibrated predictor $f:\Xs\to[0,1]^{|\Ys|}$. We assume given a trained neural network---any standard neural network for classification can be used, since they are trained to output a probability distribution over labels~\citep{guo2017calibration}.

Then, the temperature scaling approach defines a class $\Fs$ of uncertainty predictors based on $f$. Let $\phi:\Xs\to\Phi$ be the feature map of $f$ (i.e., its second-to-last layer), $\bar{f}:\Phi\to\Ys$ be the output layer of $f$, so $f=\bar{f}\circ\phi$, and
\begin{align*}
\Fs=\{T_f \bar{f} \circ \phi \mid T_f\in\mathbb{R}_+\},
\end{align*}
where $(T_f\bar{f})(z)=T_f\cdot\bar{f}(z)$. Then, the temperature scaling algorithm learns $\fh\in\Fs$ by minimizing the calibration error (\ref{eq:conv_calibration_goal}) as a function of the parameter $T_f$; this approach is depicted in \figref{fig:cal_net_1}. Intuitively, since the model family only has a single parameter, it can be estimated using very little data. 

%% learn a source-discriminator
We extend this approach to account for covariate shift. One approach would be to estimate of our bound (\ref{eq:main_bound}) using samples $(x,y)\sim p$ and $x\sim q$, and then minimize this bound over $\fh\in\Fs$ and $\gh\in\Gs$ (where $\Gs$ is to be specified). However, to simplify the problem, we separate it into two steps: (i) we train $\gh$ by minimizing the third term in (\ref{eq:main_bound}), and (ii) we train $\fh$ by minimizing the second term in (\ref{eq:main_bound}). Note that the first term of (\ref{eq:main_bound}) is constant and can be ignored.

First, we learn a source-discriminator by minimizing the third term in (\ref{eq:main_bound})---i.e., we train $\gh$ to distinguish samples $x\sim p$ from samples $x\sim q$. We do so in two steps: (i) we train an initial source-discriminator $\gh_0$, and (ii) we calibrate $\gh_0$ using temperature scaling to obtain $\gh$. In particular, consider $\Gs_0=\{\bar{g}\circ\phi\mid\bar{g}:\Phi\to[0,1]\}$, where $\phi$ is the feature map of $f$ and $\bar{g}$ is a logistic regression function, and train $\gh_0\in\Gs_0$ by minimizing the third term in (\ref{eq:main_bound}), using $\gh_0$ in place of $\gh$. Then, we calibrate $\gh_0=\bar{g}\circ\phi$ using temperature scaling---i.e., we let $\Gs=\{T_g\bar{g}\circ\phi\mid T_g \in\mathbb{R}_+\}$, where $(T\bar{g})(z)=T \cdot \bar{g}(z)$.%, and learn $\gh\in\Gs$ by minimizing the third term in (\ref{eq:main_bound}).

%% reuse a feature map
%In particular, we first train $\gh$ to be of the form $\bar{g}\circ \phi$---i.e., we reuse the feature map from the given classifier $f$. 
%% calibrate the source discriminator
%Once we have learned $\bar{g}$, we choose
%\begin{comment}
%%% note
%{\color{red}
%One practical issue for learning a source-discriminator is that minimizing the source-discriminator error in \eqnref{eq:main_bound} by gradient methods can lead exploding gradients. To avoid this issue, we minimize the following surrogate error:
%\eqas{
%	\Expop_r \[ ( s - \gh(x) )^{2 }\],
%}
%where
%$r(x \;|\; s = 1) = p(x)$, 
%$r(x \;|\; s = 0) = q(x)$, 
%and 
%$r(s=1) = r(s=0) = 0.5$, as defined in \secref{subsec:domain_discriminator}.
%\SP{justify this heuristic.}
%}
%\end{comment}

%% learn a forecaster
Second, we train $\fh\in\Fs$ by minimizing the second term in (\ref{eq:main_bound}). The overall approach is depicted in \figref{fig:cal_net_2}.
% and the learned source-dsicriminator.
%Here, we model $\bar{f} \circ \psi$ and $\bar{g}$ as a simple two layer neural network, but can be more general.
%As we discuss next, we can improve our algorithm by combining it with ideas from unsupervised domain adaptation via feature learning.

\subsection{Indistinguishable Feature Learning}
\label{subsec:algind}

Recall that we assume that the importance weights $w(x)$ are bounded, which may not hold in practice. To alleviate this issue, we leverage unsupervised domain adaptation~\citep{JMLR:v17:15-239}. In particular, our algorithm learns an \emph{auxiliary feature map} $\psi$, which is trained so that feature distribution induced by the source distribution is indistinguishable from the feature distribution induced by the target distribution (by using adversarial training).

Indistinguishable feature learning algorithms aim to learn a feature map $\phi_D:\Xs\to\Phi_D$, predictor $\bar{F}:\Phi_D\to\Ys$, and \emph{discriminator} $\bar{D}:\Phi_D\to[0,1]$ as follows: (i) $\bar{F}$ is trained to achieve good performance on predicting $\y$ given $\phi_D(x)$, (ii) $\bar{D}$ is trained to achieve good performance on predicting whether $x$ is from $p$ or $q$ given $\phi_D(x)$, and (iii) $\phi_D$ is trained so that $F=\bar{F}\circ\phi_D$ achieves good performance but $D=\bar{D}\circ\phi_D$ achieves poor performance~\citep{JMLR:v17:15-239}. Intuitively, $\phi_D$ tries to ``align'' $p$ and $q$, so that assuming $F$ achieves good performance on samples from $p$, then it achieves good performance on samples from $q$ as well.
%Theoretical bounds on the performance of this approach have been shown in prior work~\citep{ben2007analysis}.
%Next, $\bar{f}$ is unscaled top-forecaster and .
%In the case of \figref{fig:cal_net_1} and \figref{fig:cal_net_2}, $\bar{f}$ in the white box denotes that we adapt the weight from the last layer of the given classifier without any modification.
%We propose a calibration net in \figref{fig:cal_net_4}, along with its ablations \figref{fig:cal_net_2} and \figref{fig:cal_net_3}.
%One effective conventional approach for calibration, called temperature scaling \cite{guo2017calibration}, is also depicted in \figref{fig:cal_net_1}, which is also a space case of the proposed calibration net.
%% describe the calibration net
%% our method

Our key insight is that $D$ is a source-discriminator. Thus, we can use indistinguishable feature learning to train $\bar{f}=\bar{F}$ and $\bar{g}=\bar{D}$. In particular, we choose $\phi_D=\psi\circ\phi$ (where $\phi$ is the feature map of $f$), and train $\psi$, $\bar{F}$, and $\bar{D}$ using indistinguishable feature learning.
Once the indistinguishable feature map $\psi$ is trained, we fix it and re-train the source discriminator $\bar{D}$; empirically, this approach produces better importance weights than using the original source discriminator.
%a sharp IW rather that directly using the discriminator by the feature learning. 

%\begin{align}
%	\min_{F \in \Fs, D \in \Gs} \Expop_{p} \[ \| F(x) - \y \|^{2} \( \frac{1}{D(x)} - \frac{1}{2} \) \] \nonumber \\ + \lambda \Expop_{r} \[ \( D(x) - s \)^{2} \]. \label{alg:main}
%\end{align}
%We minimize this objective in two steps.
%for the calibration net \figref{fig:cal_net_4},
%(i) learn the indistinguishable feature map, and (ii) use our algorithm in Section~\ref{subsec:alg}.
%First, we train $\psi$, $\bar{g}$, and $\bar{f}$ using indistinguishable feature learning
%(\eg domain adaptation with gradient reversal layer)
%---i.e., we minimize (\ref{alg:main}) with a gradient reversal layer between $\psi$ and $\bar{g}$, but without importance weighing in the first term. Second, we use our algorithm in Section~\ref{subsec:alg} to calibrate $\bar{g}$ and $\bar{f}$, except we use the representation $\psi\circ\phi$ instead of on $\phi$---%
%---i.e., we learn a source-discriminator $\bar{g}$, and then calibrate $\bar{f}$ and $\bar{g}$ using temperature scaling parameters $T_f$ and $T_g$, respectively, by minimizing an estimate of (\ref{alg:main}) on a validation set. The key difference is that we use the auxiliary feature map $\psi$ for both $\Fs$ and $\Gs$---

Then, we train $\fh$ and $\gh$ using an approach based on the one in Section~\ref{subsec:alg}. Let $\Fs = \{T_f \bar{F} \circ \psi \circ \phi \mid T_f\in\mathbb{R}_+\}$ and $\Gs = \{T_g \bar{D} \circ \psi \circ \phi \mid T_g \in\mathbb{R}_+\}$. To train $\gh\in\Gs$, we minimize the third term of (\ref{eq:main_bound}), and to train $\fh\in\Fs$, we minimize the second term of (\ref{eq:main_bound}). We show our approach in \figref{fig:cal_net_4}. In \figref{fig:cal_net_3}, we show a variant that uses feature learning but not importance weighting.

%Note that training other calibration networks in \figref{fig:cal_nets} can be a special case of this procedure. For example, to train \figref{fig:cal_net_2}, we first train a source-discriminator $\bar{g}$ on the original feature space without training the auxiliary feature map $\psi$ on a training set. Then, we calibrate $\bar{g}$ w.r.t. $T_g$ and then use the calibrated $\bar{g}$ for importance weighted temperature scaling for $T_f$. In this case, $\bar{f}$ is the same as the last layer of the given classifier. Note that the corresponding hyper-parameters and training code of each calibration net will be publicly available.  
%\SP{add hyper-parameters and nets in the appendix if time permit.}

%% file: sections/exp.tex
%% !TEX root = ../conf_da.tex

%%
%% experiments
%%
\section{Experiments}
\label{sec:exp}

%%
%% quantitative results
%%

%% ECE under dataset shift + LeNet5: enumerate over delta
\begin{table*}[h!]
\newcolumntype{M}[1]{>{\centering\arraybackslash}m{#1}}
    \scriptsize
	\centering
	\setlength{\tabcolsep}{3pt}
	\begin{tabular}{c|c||c||c|c||c|c|c}
		% dataset name
		\makecell{data}
		%% titles
		 & \makecell{Shift} 
		 % classification error
		 & \makecell{Cls. \\ error} 
		 % baselines
		 & None 
		 & \makecell{Temp. \\ \citep{guo2017calibration}}
		 % Ours
		 & IW+Temp.
		 & \makecell{FL+Temp.}
		 & \makecell{FL+IW+Temp.}
		 \\
		 \toprule

		 %% MNIST -> MNIST
		 & {\small$\Ms \rightarrow \Ms$}	
		 % classification error
		  	& \makecell{0.89\%}
			%baselines
			& \makecell{0.83\% \\ 0.82\%}
			& \makecell{$\mathbf{0.18 \pm 0.00}$\% \\ $0.12 \pm 0.00$\%}
			%Ours
			& \makecell{${0.27 \pm 0.00}$\% \\ ${0.20 \pm 0.00}$\%}
			& \makecell{$0.24 \pm 0.04$\% \\ $\underline{0.10 \pm 0.03}$\%} 
			& \makecell{$\underline{0.21 \pm 0.02}$\% \\ $\mathbf{0.10 \pm 0.02}$\%}
		 \\
		 \cmidrule{2-8}
		 
		 %% USPS -> MNIST
		 & {\small$\Us \rightarrow \Ms$}	
		 % classification error
		  	& \makecell{47.44\%}	
		  	%baselines
			& \makecell{42.43\% \\ 42.43\%}
			& \makecell{$36.89 \pm 0.01$\% \\ $36.89 \pm 0.01$\%}
			%Ours
			& \makecell{$\mathbf{7.68 \pm 1.80}$\% \\ $\mathbf{7.47 \pm 1.97}$\%}
			& \makecell{$17.08 \pm 1.01$\% \\ $15.72 \pm 0.84$\%} 
			& \makecell{$\underline{10.58 \pm 4.67}$\% \\ $\underline{9.47 \pm 4.98}$\%}
		 \\
		\cmidrule{2-8}
		 
		 %% MNIST -> USPS
		 \multirow{-10}{*}[-5em]{\parbox[t]{1mm}{{\rotatebox[origin=c]{90}{digits}}}}
		 & {\small$\Ms \rightarrow \Us$}	
		 % classification error
			& \makecell{22.52\%}	
		  	%baselines
			& \makecell{21.26\% \\ 21.26\%}
			& \makecell{$12.16 \pm 0.01$\% \\ $11.88 \pm 0.00$\%}
			%Ours
			& \makecell{${15.92 \pm 1.38}$\% \\ ${15.86 \pm 1.41}$\%}
			& \makecell{$\underline{11.47 \pm 1.36}$\% \\ $\underline{10.80 \pm 1.00}$\%} 
			& \makecell{$\mathbf{10.85 \pm 1.45}$\% \\ $\mathbf{10.34 \pm 1.52}$\%}
		 \\
		 \cmidrule{2-8}
		 
		 %% SVHN -> MNIST
		 & {\small$\Ss \rightarrow \Ms$}
		 	% classification error
		  	& \makecell{34.14\%}	
		  	%baselines
			& \makecell{33.94\% \\ 33.94\%}
			& \makecell{$28.63 \pm 0.00$\% \\ $28.62 \pm 0.00$\%}
			%Ours
			& \makecell{${26.08 \pm 0.46}$\% \\ ${26.07 \pm 0.46}$\%}
			& \makecell{$\underline{23.37 \pm 0.42}$\% \\ $\underline{20.94 \pm 0.44}$\%} 
			& \makecell{$\mathbf{21.59 \pm 5.40}$\% \\ $\mathbf{18.87 \pm 5.40}$\%}
		 \\
		 \cmidrule{2-8}
		 
		 %% MNIST -> SVHN
		 & {\small$\Ms \rightarrow \Ss$}	
		 % classification error
		  	& \makecell{70.41\%}	
		  	%baselines
			& \makecell{59.96\% \\ 59.96\%}
			& \makecell{${10.28 \pm 0.01}$\% \\ ${10.13 \pm 0.01}$\%}
			%Ours
			& \makecell{${25.47 \pm 8.98}$\% \\ ${25.43 \pm 9.05}$\%}
			& \makecell{$\mathbf{3.90 \pm 1.96}$\% \\ $\mathbf{2.55 \pm 1.58}$\%} 
			& \makecell{$\underline{8.57 \pm 5.60}$\% \\ $\underline{7.57 \pm 5.83}$\%}
		\\
		\midrule
		
		%%
		%% office31
		
		%% Amazon -> Webcam
		 & {\small$\As \rightarrow \Ws$}	
		 % classification error
		  	& \makecell{61.67\%}
			%baselines
			& \makecell{43.82\% \\ 43.82\%}
			& \makecell{$26.69 \pm 0.01$\% \\ $26.28 \pm 0.01$\%}
			%Ours
			& \makecell{${21.82 \pm 1.20}$\% \\ ${20.36 \pm 1.50}$\%}
			& \makecell{$\underline{17.29 \pm 2.28}$\% \\ $\underline{14.70 \pm 1.80}$\%} 
			& \makecell{$\mathbf{13.49 \pm 1.62}$\% \\ $\mathbf{10.32 \pm 2.33}$\%} 
		 \\
		 \cmidrule{2-8}
		
		 %% DSLR -> Amazon
		 \multirow{-3}{*}[-1em]{\parbox[t]{1mm}{{\rotatebox[origin=c]{90}{office31}}}}
		 & {\small$\Ds \rightarrow \As$}	
		 % classification error
		  	& \makecell{70.05\%}
			%baselines
			& \makecell{35.09\% \\ 35.09\%}
			& \makecell{$66.21 \pm 0.00$\% \\ $66.21 \pm 0.00$\%}
			%Ours
			& \makecell{${65.89 \pm 0.01}$\% \\ ${65.89 \pm 0.01}$\%}
			& \makecell{$\underline{21.39 \pm 3.17}$\% \\ $\mathbf{2.53 \pm 2.83}$\%} 
			& \makecell{$\mathbf{21.25 \pm 2.84}$\% \\ $\underline{4.06 \pm 6.51}$\%} 
		 \\
		 \cmidrule{2-8}
		 
		 %% Webcam -> Amazon
		 & {\small$\Ws \rightarrow \As$}	
		 % classification error
		  	& \makecell{58.02\%}
			%baselines
			& \makecell{${27.48}$\% \\ 27.15\%}
			& \makecell{$54.59 \pm 0.00$\% \\ $54.59 \pm 0.00$\%}
			%Ours
			& \makecell{${54.20 \pm 0.01}$\% \\ ${54.20 \pm 0.01}$\%}
			& \makecell{$\underline{25.21 \pm 1.38}$\% \\ $\underline{18.31 \pm 1.66}$\%} 
			& \makecell{$\mathbf{19.85 \pm 6.54}$\% \\ $\mathbf{14.72 \pm 4.83}$\%} 
		 \\
		 \bottomrule
		
	\end{tabular}
	\vspace{-1ex}
	\caption{Empirical calibration error (ECE) on the target dataset of a neural network trained on a source dataset; we show the mean and standard deviation across 10 runs. The top number in each cell is the ECE, and the bottom number is the ECE of over-confident cases. Here, $\Ms$ denotes MNIST, $\Us$ denotes USPS, $\Ss$ denotes SVHN, $\As$ denotes Amazon, $\Ds$ denotes DSLR, and $\Ws$ denotes Webcam. ECEs with bold and underbar represent the best and the second-best results, respectively.}
	\label{tab:comp_source_network}
	\vspace{-1ex}
\end{table*}

We evaluate the effectiveness of our approach on shifts between digit classification datasets and between office object datasets. For ease of comparison, we focus on recalibrating a given classifier.

\textbf{Evaluation metric.}
%\label{subsec:eval}
%% intro the evaluation metric
%The calibration method is evaluated based on its calibration error can be derived in multiple ways based on the calibration definition \eqnref{def:calibrated}. 
We use empirical calibration error (ECE) to measure performance.
%, which is closely related to the calibration definition in (\ref{def:calibrated}), and useful for visualizing the error in a reliability diagram \citep{guo2017calibration}.
Let $f:\Xs\to\Ys$ be a label predictor, and let $\fh:\Xs\to[0,1]^{|\Ys|}$ be a forecaster that predicts the uncertainty of $f$ (\eg constructed using recalibration). First, we partition the test data $(x, \y)$ into $B$ bins (we use $B=15$) based on $\fh(x)_{f(x)}$---\ie
$\Ts_{b} = \{ (x, \y) \;|\; c_{b-1} \leq \fh(x)_{f(x)} < c_{b} \}$, 
where 
$b\in\{1,...,B\}$ and 
$0 = c_{0} \leq c_{1} \leq c_{2} \leq \dots \leq c_{B} = 1$. We denote the union of all bins as $\Ts$. 
%% intuitive explanation
Then, ECE measures the absolute calibration error across bins:
%% define metric
%\eqa{
%	\sum_{b=1}^{B} \frac{| \Xs_{b} |}{| \Ts |} 
%		\left| 
%			\frac{1}{| \Xs_{b} |} \sum_{x \in \Xs_{b}} \fh(x)_{f(x)}  - 
%			\frac{1}{| \Ys_{b} |} \sum_{\y \in \Ys_{b}} \y_{f(x)}
%		\right|, \label{eq:eval_metric}
%}
\eqa{
	\sum_{b=1}^{B} \frac{| \Ts_{b} |}{| \Ts |} 
		\left| 
			\frac{1}{| \Ts_{b} |} \sum_{(x, \y) \in \Ts_{b}} (\fh(x)_{f(x)} - \y_{f(x)})
		\right| \label{eq:eval_metric}
}
In particular, $\frac{1}{| \Ts_{b} |} \sum_{(x,\y) \in \Ts_{b}} \fh(x)_{f(x)}$ is the average predicted uncertainty for bin $b$, and $\frac{1}{| \Ts_{b} |} \sum_{(x,\y) \in \Ts_{b}} \y_{f(x)}$ is the empirical accuracy in bin $b$.

%% overconfident case
It is often worse to underestimate uncertainty than to overestimate it. Thus, we also measure the ECE restricted overconfident predicted uncertainties by considering only a one-side error in (\ref{eq:eval_metric})---\ie
\begin{align*}
	\sum_{b=1}^{B} \frac{| \Ts_{b} |}{| \Ts |} 
		\max\left\{ \! 0, \frac{1}{| \Ts_{b} |} \sum_{(x, \y) \in \Ts_{b}} (\fh(x)_{f(x)} - \y_{f(x)}) \! \right\} \!.
\end{align*}
\textbf{Digits datasets.}
We first consider shifts between MNIST \citep{lecun1989backpropagation}, USPS \citep{hull1994database}, and SVHN~\citep{netzer2011reading}.
% MNIST: 28x28, gray, 50000, 10000, 10000
MNIST is a hand-written dataset, which includes $28\times28$ gray scale images. It is split into $50,000$ training examples and $10,000$ test examples. We hold out $10,000$ validation examples used for calibration. 
% USPS: recale to 28x28, gray, (6198, 1093, 2007)
USPS is also a grayscale hand-written dataset. We rescale the dataset to $28\times28$ images so that it is compatible with classifiers trained on MNIST. It consists of a $7,291$ training examples and $2,007$ test examples, respectively. We hold out $1,093$ validation examples.
% SVHN: 32x32, splits: (62269, 10988,  26032)
Finally, SVHN (i.e., Street View House Numbers) consists of $73,257$ training examples and $26,032$ test examples, which are color images of size $32\times32$~\citep{netzer2011reading}. We hold out $10,988$ validation examples. To use SVHN images as input to a classifier trained on MNIST, we rescale them to size $28\times28$ and convert them to grayscale. 

%% office dataset
\textbf{Office31 dataset \citep{saenko2010adapting}.}
We also consider shifts between 31 office objects collected at Amazon, and captured by DSLR and webcam. We call each dataset Amazon, DSLR, and Webcam, respectively. 
All images are RGB images of various sizes; we rescale each image to $224 \times 224$. 
% Amazon: 300x300 RGB, resize to 224x224, train/val/test: 1971, 422, 424
``Amazon'' consists of $1971$ training examples, $422$ validation examples, and $424$ test examples.
% Webcam: various sized RGB, resize to 224x224, train/val/test: 556, 119, 120
``Webcam'', which consists of office images captured by a webcam, contains $556$ labeled training examples and $120$ labeled test examples; we hold out $119$ validation examples. 
% DSLR: 1000x1000 RGB, resize to 224x224, train/val/test: 348, 74, 76
``DSLR'' consists of images captured by a DSLR camera; we split them into $348$ training, $76$ test, and $74$ validation examples.

\textbf{Neural network architecture.}
%% network architecture
% forecaster feature
For the forecaster $\bar{f} \circ \psi$, we use a neural network with one hidden layer. 
% digit
When the source distribution is MNIST, USPS, or SVHN, the number of hidden units is $84$, $32$, or $256$, respectively. 
% office31
If the source distribution is Amazon, Webcam, or DSLR, the number of hidden units is 1000.
% source-discriminator
To model the source-discriminator $\bar{g}$, we also use a neural network with a single hidden layer; 
% digit
the number of hidden units are the same as the forecaster, except for the discriminator for the SVHN distribution, which has $100$ hidden units. 
% office
For the Office31 dataset, we use $1000$ hidden units for all three source distributions. 
% how to choose the hyper-parameters?
Note that we choose the network hyperparameters using the following simple heuristic, which does not depend on target labels:
choose the number of hidden neurons by starting with as many hidden neurons as there are input neurons,
and then iteratively reducing the number of neurons until the training loss converges.

\textbf{Training.}
%   
%% hyperparameters
To train each of the neural networks, we use stochastic gradient descent for $500$ epochs.
%with momentum of $0.9$ and initial learning rate of $0.001$, which decreases by a half for every $100$ epochs.
For model selection, we use standard cross validation for the source-discriminator. For the indistinguishable feature learning, we do not have target labels, so we use a simple early termination rule: terminate the feature learning when a source-discriminator loss over a training set is less than a threshold (\eg 0.2 in our experiments). Alternatively, importance weighted cross validation can be used \citep{sugiyama2007covariate}.
%we use importance weighted cross validation using Gaussian kernel density estimation with a kernel size of $100$ \cite{sugiyama2007covariate}. % instability issue
We use two approaches to avoid instability in the indistinguishable feature learning. First, we use early stopping---i.e., we stop training the indistinguishable feature map if the discriminator becomes too confident in its predicted probabilities. Second, we use dropout to regularize the discriminator. These approaches were sufficient to stabilize training of the discriminator.
To estimate the temperature scaling parameters, we use a stochastic gradient descent for $1000$ epochs on the validation set from the source distribution.
%; the initial learning rates are $0.01$ for temperature of the source discriminator $T_g$, and $2.0$ for the temperature of the forecaster $T_f$. They decay by a half every $100$ epochs. 

%% results
\textbf{Results.}
\tabref{tab:comp_source_network} summarizes the evaluation on our methods and comparing approaches. 
%% reliability diagram
In Appendix~\ref{sec:reliabilitydiagram}, we show reliability diagrams that help visualize these results~\citep{guo2017calibration}, and in Appendix~\ref{sec:runningtimeresults}, we give results on the running time of our algorithm.

\textbf{Baselines.}
We compare to two baselines---(i) the given neural network classifier $f$, which is the best trained network over a validation set, and (ii) the temperature scaling approach applied to $f$. Each cell shows the ECE (top) and ECE on over-confident cases (bottom).
To compute mean and standard deviation of ECEs, we run 10 neural network training on the same training and validation sets.
%% interpretation: Temp <-> FL+IW+Temp. (ours)
Our approach (FL+IW+Temp.) outperforms both baselines. This result is due to the fact that our approach leverages unlabeled examples from the target distribution to improve the estimated calibration errors. Note that the over-confident ECE also improves.
%% M->M identity
One desirable property of our approaches is that its uncertainty predictions remain good even if there is no shift in the dataset---e.g., in the case of the shift from MNIST to MNIST. Finally, the ablation FL+Temp. can be thought of as the baseline of using invariant feature learning for domain adaptation~\citep{JMLR:v17:15-239} in conjunction with temperature scaling to obtain calibrated probabilities.

%% interpretation: (ablation) No FL (i.e., IW+Temp <-> FL+IW+Temp)
\textbf{Ablations.}
We compare to two ablations: one without feature learning and one without importance weighting. As can be seen, feature learning in general improves ECE.
%This means the feature learning could help to enforce the bounded importance weight assumption, as expected.
For the shift from USPS to MNIST, importance weighting without feature learning produces a better result, most likely since this shift is small.
% U2M: FL original weightdecay = 5e-3
% weight decay = 1e-3, ECE =  8.28 +- 4.34, 
% weight decay = 1e-4, ECE = 12.43 +- 3.14
%Further experiments reveal that a different hyper-parameter for the feature learning leads a different ECE. 
We may be able to improve the performance of feature learning in this case by tuning hyperparameters.
%When we apply weight decay of $10^{-3}$, $5\times10^{-3}$, and $10^{-4}$ during feature learning, the corresponding ECEs are $8.28\% \pm 4.34$, $10.58\% \pm 4.67$, and $12.43\% \pm 3.14$, respectively. This implies a better way of choosing hyper-parameters is required without target labels or by exploiting a little target labels. 
%We believe this result may be because the indistinguishable feature map $\psi\circ\phi$ is too different than the original feature map $\phi$. Thus, the calibration according to $\psi\circ\phi$ may no longer reflect the calibration according to $\phi$ alone.

%% summary
Next, importance weighting substantially improves ECE. The largest gains are when the target distribution is similar to but more varied than the target distribution (e.g., USPS to MNIST). In these cases, importance weighting accounts for the uncertainty due to the increased variance of the target distribution.

%\SP{{\color{red}Mention on the comparison with UDA?}}

%% interpretation: high variance of ECE
\begin{figure}
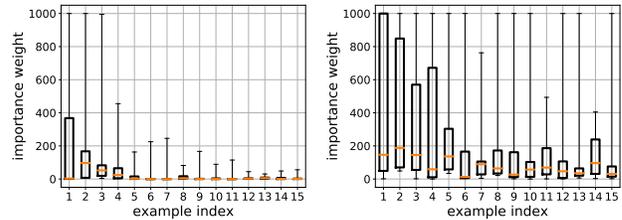

\centering
	\begin{subfigure}[b]{0.492\linewidth}
	\centering
	\includegraphics[width=\linewidth]{{{figs/IW_analysis/MNIST2SVHN/plot_iws_box.png}}}
	\caption{$\Ms \rightarrow \Ss$ (ECE std. $5.60$)}
	\label{fig:IW_analysis_a}
	\end{subfigure}
	\begin{subfigure}[b]{0.492\linewidth}
	\centering
	\includegraphics[width=\linewidth]{{{figs/IW_analysis/MNIST2USPS/plot_iws_box.png}}}
	\caption{$\Ms \rightarrow \Us$ (ECE std. $1.45$)}
	\label{fig:IW_analysis_b}
	\end{subfigure}
	\caption{These plots show the distribution of our importance weight estimates. For each source example, we estimate the importance weights on 10 runs. We choose 15 examples with the largest corresponding average estimated importance weight. For these 15 examples, we plot the median, minimum, and maximum of the estimated importance weights across the 10 runs.}
	\label{fig:IW_analysis}
\end{figure}

\textbf{ECE variance.}
%
%% description
Some of our benchmarks exhibit high variance ECEs---in particular, the benchmark MNIST to SVHN. Note that this variance is caused by randomness in our algorithm, since the randomness across runs is only based on different random choices by our algorithm. In particular, we believe the variance is caused by the variance in our learned source-discriminator $\gh$, most likely due to optimization error when fitting $\bar{g}$ and $\psi$. This variance can cause our importance weights to vary drastically across runs. For example, \figref{fig:IW_analysis} shows the distribution of the estimated importance weights on the benchmarks MNIST to SVHN (which has high ECE variance) and MNIST to USPS (which has low ECE variance) over 10 runs. As can be seen, the the importance weights vary greatly across runs.

Interestingly, the variances appear to be larger on the shift from MNIST to USPS, even though the variance for the ECE on this benchmark is smaller. However, note that for the shift from MNIST to SVHN, the importance weights are close to zero for most examples. We believe that for the shift from MNIST to USPS, even if the importance weights have high variance, the noise is averaged out across many examples. In contrast, for SVHN, there are very few relevant examples; thus, if we obtain poor estimates of the importance weights for these relevant examples, then the subsequent calibration step will perform poorly.

%% file: sections/disc.tex
\section{Conclusion}

We have proposed a novel approach for calibrated prediction in the presence of covariate shift, and empirically demonstrated that our approach outperforms existing ones. Future work includes handling the case of online covariate shift (\ie the classifier is observing new examples sequentially), and extending our results beyond the classification setting.

%\SP{cite and attach CAM paper \cite{cam19draft}}

%% file: sections/appendix.tex
%% !TEX root = ../conf_da.tex

\clearpage
\clearpage
%%
%% Appendix
%% 
\onecolumn
\appendix

\section{Additional Results}
\label{sec:additionalresults}

\subsection{Reliability Diagrams}
\label{sec:reliabilitydiagram}

%% describe

The \emph{reliability diagram} is an intuitive visualization of the empirical calibration error (ECE) in (\ref{eq:eval_metric})  \citep{degroot1983comparison,niculescu2005predicting}. In particular, the diagram shows the averaged predicted uncertainty (\ie $\frac{1}{| \Ts_{b} |} \sum_{(x, \y) \in \Ts_{b}} \fh(x)_{f(x)}$) on the $x$-axis, and the empirical accuracy (\ie $\frac{1}{| \Ts_{b} |} \sum_{(x, \y) \in \Ts_{b}} \y_{f(x)}$) on the $y$-axis. Thus, if bars in the reliability diagram aligns with the diagonal, the ECE of the forecaster in consideration is zero. If the bars are below the diagonal, then the forecaster is over-confident on its uncertainty predictions. Along with the accuracy and confidence plots, each bar is weighted by the fraction of examples in each bin $b$ (\ie $\frac{|\Ts_b|}{|\Ts|}$), which is also reflected in the ECE in (\ref{eq:eval_metric}). 

We compare the reliability diagram for the temperature scaling approach and for our approach in 
\figref{fig:M2U_rel}, \figref{fig:U2M_rel}, \figref{fig:S2M_rel}, \figref{fig:M2S_rel}, 
\figref{fig:A2W_rel}, \figref{fig:D2A_rel}, and \figref{fig:W2A_rel}.

\subsection{Running Time}
\label{sec:runningtimeresults}

As with existing approaches to calibrated prediction~\citep{guo2017calibration}, our approach relies on a second phase of training to calibrate the predicted probabilities. We measure the overhead of our calibration step compared to the time used to train the indistinguishable feature map. The results are:
\begin{itemize}
% \item $\Ds \rightarrow \As$: 9015 sec (our overhead) vs. 18591 sec (total)
% \item $\Ms \rightarrow \Ss$: 1460 sec (our overhead) vs. 5897 sec (total)
\item $\Ms \rightarrow \Ms$: 
7840.8589 sec. (our overhead) vs. 
1324.9306 sec. (total)

\item $\Us \rightarrow \Ms$: 
14939.4707 sec. (our overhead) vs. 
1001.3495 sec. (total)

\item $\Ms \rightarrow \Us$: 
16900.7378 sec. (our overhead) vs. 
1217.413 sec. (total)

\item $\Ss \rightarrow \Ms$: 
16437.2544 sec. (our overhead) vs. 
581.6673 sec. (total)

\item $\Ms \rightarrow \Ss$: 
4437.1197  sec. (our overhead) vs. 
1460.2213 sec. (total)

\item $\As \rightarrow \Ws$: 
16068.9053 sec. (our overhead) vs. 
907.5134 sec. (total)

\item $\Ds \rightarrow \As$: 
9576.0645 sec. (our overhead) vs. 
9015.7301 sec. (total)

\item $\Ws \rightarrow \As$: 
18602.4316 sec. (our overhead) vs. 
7217.6859 sec. (total)

\end{itemize}
In all but one case, the overhead from calibration is less than $1/4$ the total time taken (\ie for both calibration and indistinguishable feature learning). This overhead is reasonable for obtaining calibrated probabilities. 
%We will add detailed results to our paper. 
%\SP{{\color{red}how much details do we want to add?}}

%% M->U
\begin{figure}[h]
\centering
\begin{subfigure}[b]{0.32\linewidth}
	\centering
	\includegraphics[width=\linewidth]{{{figs/plots_rel_diag/MNIST2USPS/exp_Naive_9/rel_diag.png_conf_acc}}}
	\includegraphics[width=\linewidth]{{{figs/plots_rel_diag/MNIST2USPS/exp_Naive_9/rel_diag.png_conf_acc_freq}}}
   	\caption{None}
\end{subfigure}
\begin{subfigure}[b]{0.32\linewidth}
	\centering
	\includegraphics[width=\linewidth]{{{figs/plots_rel_diag/MNIST2USPS/exp_Temp_9/rel_diag.png_conf_acc}}}
	\includegraphics[width=\linewidth]{{{figs/plots_rel_diag/MNIST2USPS/exp_Temp_9/rel_diag.png_conf_acc_freq}}}
   	\caption{Temp. scaling}
\end{subfigure}
\begin{subfigure}[b]{0.32\linewidth}
	\centering
	\includegraphics[width=\linewidth]{{{figs/plots_rel_diag/MNIST2USPS/exp_Temp_FL_IW_9/rel_diag.png_conf_acc}}}
	\includegraphics[width=\linewidth]{{{figs/plots_rel_diag/MNIST2USPS/exp_Temp_FL_IW_9/rel_diag.png_conf_acc_freq}}}
   	\caption{Temp.+FL+IW}
\end{subfigure}
\caption{Reliability diagram of the shift $\Ms \rightarrow \Us$ from one experiment among ten.}
\label{fig:M2U_rel}
\end{figure}

%% U->M
\begin{figure}[h]
\centering
\begin{subfigure}[b]{0.32\linewidth}
	\centering
	\includegraphics[width=\linewidth]{{{figs/plots_rel_diag/USPS2MNIST/exp_Naive_9/rel_diag.png_conf_acc}}}
	\includegraphics[width=\linewidth]{{{figs/plots_rel_diag/USPS2MNIST/exp_Naive_9/rel_diag.png_conf_acc_freq}}}
   	\caption{None}
\end{subfigure}
\begin{subfigure}[b]{0.32\linewidth}
	\centering
	\includegraphics[width=\linewidth]{{{figs/plots_rel_diag/USPS2MNIST/exp_Temp_9/rel_diag.png_conf_acc}}}
	\includegraphics[width=\linewidth]{{{figs/plots_rel_diag/USPS2MNIST/exp_Temp_9/rel_diag.png_conf_acc_freq}}}
   	\caption{Temp. scaling}
\end{subfigure}
\begin{subfigure}[b]{0.32\linewidth}
	\centering
	\includegraphics[width=\linewidth]{{{figs/plots_rel_diag/USPS2MNIST/exp_Temp_FL_IW_9/rel_diag.png_conf_acc}}}
	\includegraphics[width=\linewidth]{{{figs/plots_rel_diag/USPS2MNIST/exp_Temp_FL_IW_9/rel_diag.png_conf_acc_freq}}}
   	\caption{Temp.+FL+IW}
\end{subfigure}
\caption{Reliability diagram of the shift $\Us \rightarrow \Ms$ from one experiment among ten.}
\label{fig:U2M_rel}
\end{figure}

%% S->M
\begin{figure}[h]
\centering
\begin{subfigure}[b]{0.32\linewidth}
	\centering
	\includegraphics[width=\linewidth]{{{figs/plots_rel_diag/SVHN2MNIST/exp_Naive_0/rel_diag.png_conf_acc}}}
	\includegraphics[width=\linewidth]{{{figs/plots_rel_diag/SVHN2MNIST/exp_Naive_0/rel_diag.png_conf_acc_freq}}}
   	\caption{None}
\end{subfigure}
\begin{subfigure}[b]{0.32\linewidth}
	\centering
	\includegraphics[width=\linewidth]{{{figs/plots_rel_diag/SVHN2MNIST/exp_Temp_0/rel_diag.png_conf_acc}}}
	\includegraphics[width=\linewidth]{{{figs/plots_rel_diag/SVHN2MNIST/exp_Temp_0/rel_diag.png_conf_acc_freq}}}
   	\caption{Temp. scaling}
\end{subfigure}
\begin{subfigure}[b]{0.32\linewidth}
	\centering
	\includegraphics[width=\linewidth]{{{figs/plots_rel_diag/SVHN2MNIST/exp_Temp_FL_IW_0/rel_diag.png_conf_acc}}}
	\includegraphics[width=\linewidth]{{{figs/plots_rel_diag/SVHN2MNIST/exp_Temp_FL_IW_0/rel_diag.png_conf_acc_freq}}}
   	\caption{Temp.+FL+IW}
\end{subfigure}
\caption{Reliability diagram of the shift $\Ss \rightarrow \Ms$ from one experiment among ten.}
\label{fig:S2M_rel}
\end{figure}

%% M->S
\begin{figure}[h]
\centering
\begin{subfigure}[b]{0.32\linewidth}
	\centering
	\includegraphics[width=\linewidth]{{{figs/plots_rel_diag/MNIST2SVHN/exp_Naive_9/rel_diag.png_conf_acc}}}
	\includegraphics[width=\linewidth]{{{figs/plots_rel_diag/MNIST2SVHN/exp_Naive_9/rel_diag.png_conf_acc_freq}}}
   	\caption{None}
\end{subfigure}
\begin{subfigure}[b]{0.32\linewidth}
	\centering
	\includegraphics[width=\linewidth]{{{figs/plots_rel_diag/MNIST2SVHN/exp_Temp_9/rel_diag.png_conf_acc}}}
	\includegraphics[width=\linewidth]{{{figs/plots_rel_diag/MNIST2SVHN/exp_Temp_9/rel_diag.png_conf_acc_freq}}}
   	\caption{Temp. scaling}
\end{subfigure}
\begin{subfigure}[b]{0.32\linewidth}
	\centering
	\includegraphics[width=\linewidth]{{{figs/plots_rel_diag/MNIST2SVHN/exp_Temp_FL_IW_9/rel_diag.png_conf_acc}}}
	\includegraphics[width=\linewidth]{{{figs/plots_rel_diag/MNIST2SVHN/exp_Temp_FL_IW_9/rel_diag.png_conf_acc_freq}}}
   	\caption{Temp.+FL+IW}
\end{subfigure}
\caption{Reliability diagram of the shift $\Ms \rightarrow \Ss$ from one experiment among ten.}
\label{fig:M2S_rel}
\end{figure}

%% A->W
\begin{figure}[h]
\centering
\begin{subfigure}[b]{0.32\linewidth}
	\centering
	\includegraphics[width=\linewidth]{{{figs/plots_rel_diag/Amazon2Webcam/exp_Naive_9/rel_diag.png_conf_acc}}}
	\includegraphics[width=\linewidth]{{{figs/plots_rel_diag/Amazon2Webcam/exp_Naive_9/rel_diag.png_conf_acc_freq}}}
   	\caption{None}
\end{subfigure}
\begin{subfigure}[b]{0.32\linewidth}
	\centering
	\includegraphics[width=\linewidth]{{{figs/plots_rel_diag/Amazon2Webcam/exp_Temp_9/rel_diag.png_conf_acc}}}
	\includegraphics[width=\linewidth]{{{figs/plots_rel_diag/Amazon2Webcam/exp_Temp_9/rel_diag.png_conf_acc_freq}}}
   	\caption{Temp. scaling}
\end{subfigure}
\begin{subfigure}[b]{0.32\linewidth}
	\centering
	\includegraphics[width=\linewidth]{{{figs/plots_rel_diag/Amazon2Webcam/exp_Temp_FL_IW_9/rel_diag.png_conf_acc}}}
	\includegraphics[width=\linewidth]{{{figs/plots_rel_diag/Amazon2Webcam/exp_Temp_FL_IW_9/rel_diag.png_conf_acc_freq}}}
   	\caption{Temp.+FL+IW}
\end{subfigure}
\caption{Reliability diagram of the shift $\As \rightarrow \Ws$ from one experiment among ten.}
\label{fig:A2W_rel}
\end{figure}

%% D->A
\begin{figure}[h]
\centering
\begin{subfigure}[b]{0.32\linewidth}
	\centering
	\includegraphics[width=\linewidth]{{{figs/plots_rel_diag/DSLR2Amazon/exp_Naive_9/rel_diag.png_conf_acc}}}
	\includegraphics[width=\linewidth]{{{figs/plots_rel_diag/DSLR2Amazon/exp_Naive_9/rel_diag.png_conf_acc_freq}}}
   	\caption{None}
\end{subfigure}
\begin{subfigure}[b]{0.32\linewidth}
	\centering
	\includegraphics[width=\linewidth]{{{figs/plots_rel_diag/DSLR2Amazon/exp_Temp_9/rel_diag.png_conf_acc}}}
	\includegraphics[width=\linewidth]{{{figs/plots_rel_diag/DSLR2Amazon/exp_Temp_9/rel_diag.png_conf_acc_freq}}}
   	\caption{Temp. scaling}
\end{subfigure}
\begin{subfigure}[b]{0.32\linewidth}
	\centering
	\includegraphics[width=\linewidth]{{{figs/plots_rel_diag/DSLR2Amazon/exp_Temp_FL_IW_9/rel_diag.png_conf_acc}}}
	\includegraphics[width=\linewidth]{{{figs/plots_rel_diag/DSLR2Amazon/exp_Temp_FL_IW_9/rel_diag.png_conf_acc_freq}}}
   	\caption{Temp.+FL+IW}
\end{subfigure}
\caption{Reliability diagram of the shift $\Ds \rightarrow \As$ from one experiment among ten.}
\label{fig:D2A_rel}
\end{figure}

%% W->A
\begin{figure}[h]
\centering
\begin{subfigure}[b]{0.32\linewidth}
	\centering
	\includegraphics[width=\linewidth]{{{figs/plots_rel_diag/Webcam2Amazon/exp_Naive_1/rel_diag.png_conf_acc}}}
	\includegraphics[width=\linewidth]{{{figs/plots_rel_diag/Webcam2Amazon/exp_Naive_1/rel_diag.png_conf_acc_freq}}}
   	\caption{None}
\end{subfigure}
\begin{subfigure}[b]{0.32\linewidth}
	\centering
	\includegraphics[width=\linewidth]{{{figs/plots_rel_diag/Webcam2Amazon/exp_Temp_1/rel_diag.png_conf_acc}}}
	\includegraphics[width=\linewidth]{{{figs/plots_rel_diag/Webcam2Amazon/exp_Temp_1/rel_diag.png_conf_acc_freq}}}
   	\caption{Temp. scaling}
\end{subfigure}
\begin{subfigure}[b]{0.32\linewidth}
	\centering
	\includegraphics[width=\linewidth]{{{figs/plots_rel_diag/Webcam2Amazon/exp_Temp_FL_IW_1/rel_diag.png_conf_acc}}}
	\includegraphics[width=\linewidth]{{{figs/plots_rel_diag/Webcam2Amazon/exp_Temp_FL_IW_1/rel_diag.png_conf_acc_freq}}}
   	\caption{Temp.+FL+IW}
\end{subfigure}
\caption{Reliability diagram of the shift $\Ws \rightarrow \As$ from one experiment among ten.}
\label{fig:W2A_rel}
\end{figure}